# Deep Synthetic Minority Over-Sampling Technique

Hadi Mansourifar  
University of Houston

Weidong Shi  
University of Houston

**Abstract**

Synthetic Minority Over-sampling Technique (SMOTE) is the most popular over-sampling method. However, its random nature makes the synthesized data and even imbalanced classification results unstable. It means that in case of running SMOTE *n* different times, *n* different synthesized instances are obtained with *n* different classification results. To address this problem, we adapt the SMOTE idea in deep learning architecture. In this method, a deep neural network regression model is used to train the inputs and outputs of traditional SMOTE. Inputs of the proposed deep regression model are two randomly chosen data points which are concatenated to form a double size vector. The outputs of this model are corresponding randomly interpolated data points between two randomly chosen vectors with original dimension. The experimental results show that, Deep SMOTE can outperform traditional SMOTE in terms of precision, F1 score and Area Under Curve (AUC) in majority of test cases.

## 1  Introduction

One of the most challenging problems in binary classification is dealing with imbalanced datasets in which the class distribution is skewed because the positive instances significantly outnumber the negative instances. Researchers deal with the class imbalanced problem in many real-world applications, such as diabetes detection [1], breast cancer diagnosis and survival prediction [2,3], Parkinson diagnosis [4], bankruptcy prediction [5], credit card fraud detection [6] and default probability prediction [7]. In these applications, the main task is to detect a minority instance. However, standard classifiers are generally inefficient due to low rate occurrence of the minority instances. A standard classifier trains the models with bias toward the majority class which leads to high overall accuracy and poor recall score since a large fraction of minority instances would be labeled as majority instance. Solutions to address the class imbalanced problem fall into two categories: data driven approaches and algorithmic approaches. Data driven techniques [8] aim to balance the class distributions of a dataset before feeding the output into a classification algorithm by either over-sampling or undersampling the data. When the dataset is highly imbalanced, under-sampling could lead to significant loss of information. In such cases, over-sampling has proven to be more effective for dealing with class-imbalanced problem. Chawla, Bowyer, Hall, and Kegelmeyer [9] proposed an over-sampling technique called Synthetic Minority Over-sampling Technique (SMOTE). It interpolates synthetic instances along a line segment which connects two randomly chosen data points. SMOTE is the most popular over-sampling method due to its simplicity, computational efficiency, and superior performance [10]. However, SMOTE blindly synthesizes new data in minority class without considering the majority instances, especially in vicinity regions with majority class. On the other hand, the common problem of SMOTE variations is non-stable results due to their random nature meaning that a unique set of synthesized data and classification results are not guaranteed. In fact, in case of running SMOTE *n* different times, *n* different synthesized instances are obtained with *n* different classification results. In this paper, we adopt SMOTE idea in a deep learning architecture called Deep SMOTE to make synthesized data more stable and classification results more efficient. Deep SMOTE is based on a deep neural network regression model to train the inputs and outputs of traditional SMOTE. Inputs of this deep regression model are two randomly chosen data points which are concatenated to form a double size vector. The model is trained to return the double sized input vector to the original dimension using a randomly interpolated data point between two corresponding concatenated data points. Concatenation of two randomly chosen data points to form a double size input vector increases the size of training instances from *n* to *C(n,2)* which is vital to train a deep regression model. Besides, a trained model is more stable tool to synthesize data. The experimental results show that, Deep SMOTE can outperform traditional SMOTE in terms of F1 and AUC scores in the majority of test cases. Our contributions are as follows.

- We propose a novel approach to adopt SMOTE idea in deep neural network architecture to synthesize more efficient instances.
- We propose Deep Adversarial SMOTE (DASMOTE) by unsupervised training of Deep SMOTE model in adversarial mode.

- We use vector concatenation to increase the minority training instances from *n* to *C(n,2)*.

- Our experimental results show the superiority of Deep SMOTE and DA-SMOTE versus SMOTE in majority of cases in terms of F1 and AUC scores.

The rest of the paper is organized as follows. Section 2 presents the related works. Section 3 demonstrates the Deep SMOTE. Section 4 introduces DA-SMOTE. Section 5 and section 6 present the experimental results and finally section 7 concludes the paper.

## 2 Related Works

Re-sampling methods to tackle the imbalanced data classification fall into three categories: Over-sampling, under-sampling and hybrid methods.

**2.1 Over-sampling methods** SMOTE [9] is the most popular over-sampling method due to its simplicity, computational efficiency, and superior performance [10]. However, SMOTE blindly synthesizes new data in minority class without considering the majority instances, especially in vicinity regions with majority class. To address this problem, Han et al. proposed Borderline-SMOTE [11], which focuses only on borderline instances in the majority class vicinity regions. However, the precision rate can be highly impacted because classifier fails to detect instances belonging to majority class. Although a superior oversampling method should ideally improve the minority class detection rate, it must not lead to disability to detect majority instances. To solve this problem, Barua et al. [12] proposed MWMOTE, a two-step weighted approach that extends Borderline-SMOTE and ADASYN using the information of the majority instances that lie close to the borderline. Also, Bunkhumpornpat et al. [13] proposed DBSMOTE which uses DBSCAN to evaluate the density of each region and then over-samples inside each region to avoid synthesizing an instance inside majority class. A-SUWO [14] is also a clustering-based method designed to identify groups of minority samples that are not overlapped with clusters from the majority class. However, it underestimates the role of noise or mislabeled datapoints which makes it hard to find non-overlapping regions. To Address this problem, Ma et al. [15] proposed denoising and removing outliers before over-sampling.

**2.2 GAN based data augmentation** Generative adversarial nets (GANs) [16] and variational autoencoders [31] have proven to be good at solving many tasks. In recent years, GANs have been used successfully at data augmentation [17,18, 19]. The main idea is to train a generative network in adversarial mode against a discriminator network. Since the invention of GAN, it has been well used in different machine learning applications [20,21,22,33,34], especially in computer vision and image processing.

**2.3 Under-sampling methods** Kubat and Matwin [23] proposed the first under-sampling method by electing the majority instances while keeping the original population of the minority instances unchanged. As a performance measure, the geometric mean was used, which is interpreted as a single point on the ROC curve. Kubat [24] proposed the SHRINK to detect the best possible regions in majority class for under-sampling. It searches for the "best positive region" in overlapping regions of minority (positive) and majority (negative) instances. Kang et al.[27] proposed a new under-sampling scheme by incorporating a noise filter before executing re-sampling. They discussed that, noisy minority examples may reduce the performance of classifiers. They concluded that denoising the data help under-sampling methods to train more efficient classifiers. Dal Pozzolo et al.[28] formalized the relationship between conditional probability in the presence and absence of under-sampling and how under- sampling affects the posterior probability of a machine learning model. Lin et al. [29] used two clustering techniques as data preprocessing step, considering the number of clusters in the majority class to be equal to the number of data points in the minority class.

**2.4 Hybrid Methods** Ling et al. [25] proposed combination of over-sampling of the minority class with under-sampling of the majority class. In this research, test examples were ranked by a confidence measure and then lift analysis was used instead of accuracy to measure a classifier's performance. Solberg et al. [26] used over-sampling to reach 100 data samples from the oil slick, and random under-sampling was used to sample 100 samples from the non-oil slick class (majority) to create a new dataset with balance class distribution. Junsomboon et al. [30] proposed a hybrid solution by combining Neighbor Cleaning Rule (NCL) and Synthetic SMOTE to reduce diagnostic mistake in medical systems.

## 3 Deep SMOTE

We propose an over-sampling approach called Deep SMOTE in which the minority class is over-sampled by a neural network model. This approach is inspired by traditional SMOTE which has been proven to be successful in many applications. In proposed method, the minority class is over-sampled by a deep model which has been trained to receive two randomly chosen minority instances as input and synthesizes a novel data point in original dimension along the line segment joining the inputs. To train a neural network to accept two different data points and generate an interpolated data points in between, we concatenate the inputs to form a double size vector as shown in Figure 1. Deep SMOTE is formulized as follows. Given a set of minority instances $(x_1, x_2, \cdots, x_w)$ where $x_i \in R_n$, $m$ training data points are created as $(x'_1, y_1), (x'_2, y_2), \cdots, (x'_m, y_m)$ where $x'_i \in R^{2n}$, $y_i \in R^n$ by concatenation of $x_s$ and $x_t$ where $s$, $t$ are two randomly chosen numbers, $1 < s < w, 1 < t < w$ and $y_i$ is an interpolated data point along the line segment joining $x_s$, $x_t$. Figure 1 shows the architecture of Deep SMOTE model. Training a model to interpolate a data point given two randomly chosen data points effectively forces the decision region of the minority class to become more general since the model is trained based on most frequently observed training instances. Here, we summarize the similarities between traditional SMOTE and Deep SMOTE as mentioned in previous section.

• In both methods, local minority neighborhoods are chosen randomly.

• In both methods, target pair of instances are chosen randomly from local neighborhood.

• interpolation plays an important role in both methods.

Algorithm 1,2 provide required steps for Deep SMOTE training and over-sampling.

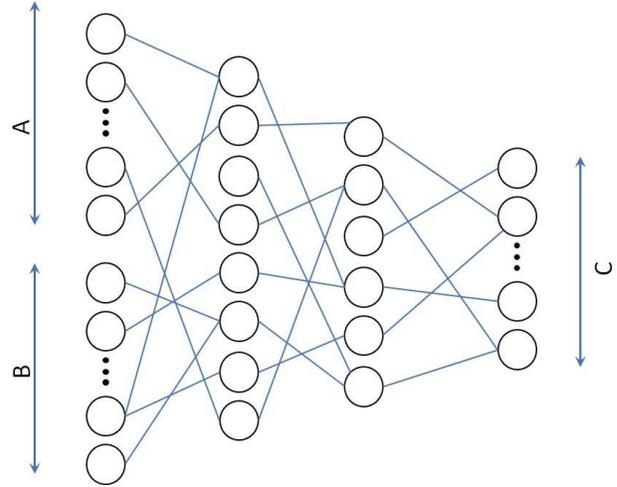

Figure 1: The architecture of Deep SMOTE. Vector A is randomly chosen data point 1 and Vector B is randomly chosen data point 2 and Vector C is an interpolated data points between vector A,B where Size(C)=Size(A)=Size(B).

After training a model, it is used to over-sample the minority class to create balance class distribution.

---

**Algorithm 1** Deep SMOTE Training (**X, T, E, L**)
**Input**: Minority instances **X**; Number of training instances **T**; Epoch Number **E**; Number of hidden nodes at each layer **L**
**Output**: Model **M**

---

1: Given **X** randomly choose **T** different data point pairs from **X** and save them in **U**.
2: Given **U**, Form **X´** by concatenation of the pairs.
3: Given **U**, form **Y´** by interpolating a new data point on connecting line of each pair.
4: Given **X´, Y´, E, L** train a neural regression model **M**.
5: Return **M**.

---

**Algorithm 2** Deep SMOTE Over-Sampling (**X, M, D**)
**Input**: Minority instances **X**; Trained model **M**;
**D**=Number of majority instances - Number of minority instances

---

1: Given **X** as minority instances, randomly choose **D** data pairs from **X** and save in **U**.
2: Given **U**, Form **X´** by concatenation of the pairs.
3: Given **M** and **U** predict the results and save in **Y´´**.
4: Merge **X** and **Y´´** and save the results in **O**.
5: Return **O**.

---

Here, we summarize the differences between traditional SMOTE and Deep SMOTE.

• In Deep SMOTE interpolation is used to generate outputs of a neural network regression model. The trained model is used eventually to synthesize new data for minority class. However, SMOTE uses interpolation to synthesize new data.

• Despite traditional SMOTE, vector concatenation is used in Deep SMOTE.

## 4 Deep Adversarial Synthetic Over-Sampling Technique ( DA-SMOTE)

DA-SMOTE is a novel over-sampling method inspired by three different ideas including SMOTE, GANs and Deep SMOTE. DA-SMOTE works based on training a neural network regression model in adversarial mode. The main difference between DA-SMOTE and Deep SMOTE is that DA-SMOTE doesn't need interpolation to train the regression model. In DA-SMOTE, a generator competes with a discriminator to achieve the best possible weights to transfer the double size vectors into a point between them with original dimension. In fact, training the generator in adversarial mode helps us to find the best data point between two concatenated data points instead of random guess for interpolating a synthesized data point. In both methods, local minority neighborhoods and target pair of instances are chosen randomly from local neighborhoods. The training algorithm of DA-SMOTE is very similar to GAN training algorithm as follows.

---

**Algorithm 3** Minibatch stochastic gradient descent training of Deep Adversarial SMOTE.

1: **for** number of training iterations **do**
2:   **for** k steps **do**
3:     Sample minibatch of $m$ data pairs $\{z^{(1)},...,z^{(m)}\}$ from minority prior and concatenate each selected pairs $p_g(z)$.
4:     Sample minibatch of $m$ sample $\{x^{(1)},...,x^{(m)}\}$ from data generating distribution $p_{data}(x)$.
5:     Update the discriminator by ascending its stochastic gradient.
$$\Delta_{\theta_d} \frac{1}{m} \sum_{i=0}^{m} [logD(x^{(i)})] + [log(1 - D(G(z^{(i)})))]$$
6:   **end for**
7:   Sample minibatch of $m$ minority data pairs $\{z^{(1)},...,z^{(m)}\}$ from minority prior and concatenate each selected pair $p_g(z)$.
8:   $\Delta_{\theta_g} \frac{1}{m} \sum_{i=0}^{m} [log(1 - D(G(z^{(i)})))]$
9: **end for**

---

Here, we summarize similarities between DASMOTE and Deep SMOTE:

- The goal of both is to over-sample the minority instances.
- In both methods, local minority neighborhoods are chosen randomly.
- In both methods, target pair of instances are chosen randomly from local neighborhoods.

Also, differences between traditional SMOTE, Deep SMOTE and DA-SMOTE are as follows. • Despite SMOTE, interpolation doesn't play a role in DA-SMOTE.

- Despite Deep SMOTE, DA-SMOTE is unsupervised meaning that it doesn't need interpolation results to train the model. • Despite traditional SMOTE, vector concatenation is used in DA-SMOTE.

Since DA-SMOTE works in adversarial mode, its algorithm is very similar to the GANs. One of the main differences of DA-SMOTE and GANs is the latent space.

- **Input type :** The latent vector in the GANs is a noise vector randomly selected in a specified range. However, the latent vector in DA-SMOTE is formed by two concatenated minority instances.
- **Input dimensionality:** There is no any limitation for latent vector dimension in GANs. However, the latent vector dimensionality in DA-SMOTE is limited to double size of original data dimension.

Another difference between GANs and DA-SMOTE is that, available training size of DA-SMOTE is significantly larger than GANs because vector concatenation lets us to increase the size of training data from $n$ to $C(n,2)$. We first prove that, vector concatenation can significantly increase the efficiency of Deep SMOTE and DA-SMOTE models by increasing the size of training data. The reason is that, the nearest neighbor of $X$ converges almost surely to $X$ as the training size grows to infinity [32].

**Lemma 1.** *(Theorem Convergence of Nearest Neighbor)* If $X_1, X_2, ..., X_n$ are i.i.d. in a separable metric space $\chi$, $X'_n(X) \triangleq X'_n$ is the closest of the $X_1, X_2, ..., X_n$ to $X$ in a metric $D(\cdot, \cdot)$, then
$$X'_n \to X \quad a.s.$$
*Proof.* Let $B_r(x)$ be the (closed) ball of radius r centered at x

$$(4.1) \qquad B_r(x) \triangleq \{z \in \mathbb{R}^d : D(z,x) \leq r\}$$

for any $r > 0$

$$(4.2) \quad P(B_r(x)) \triangleq P_r[z \in B_r(x)] = \int_{B_r(x)} p(z)d(z) > 0$$

Then, for any $\delta > 0$, we have

$$(4.3) \quad P_r\left[\min_{i=0,1,2,...} \{D(x_i,x) \geq \delta\}\right] = \left[1 - P(B_r(x))\right]^n \to 0$$

There exists a rational point $a_{\bar{x}}$ such that $a_{\bar{x}} \in B_{\bar{r}/3}(\bar{x})$. Where, for some $\bar{r}$, we have $P(B_{\bar{r}}(\bar{x})) = 0$

Table 1: Tested datasets and their characteristics.

| Name | Instance | Attribute | Pos | Neg | Minority |
|---|---|---|---|---|---|
| Pima | 768 | 8 | 268 | 500 | 34.9% |
| WBC | 699 | 9 | 241 | 458 | 34.5% |
| Haberman | 306 | 3 | 81 | 225 | 26.47% |
| Ionosphere | 351 | 34 | 126 | 225 | 35.89% |
| Parkinson | 195 | 23 | 48 | 147 | 32.65% |
| Blood | 748 | 4 | 178 | 570 | 23.79% |
| Bankruptcy-1 | 7027 | 64 | 271 | 6756 | 3.85% |
| Bankruptcy-2 | 10173 | 64 | 400 | 9773 | 3.93% |
| Bankruptcy-3 | 10503 | 64 | 495 | 10008 | 4.71% |
| Bankruptcy-5 | 5910 | 64 | 410 | 5500 | 6.93% |

Consequently, there exists a small sphere $B_{\bar{r}/2}(\bar{x})$ such that

$$(4.4) \quad B_{\bar{r}/2}(\bar{x}) \subset B_{\bar{r}}(\bar{x}) \Rightarrow P(B_{\bar{r}/2}(\bar{x})) = 0$$

Also, $\bar{x} \in B_{\bar{r}/2}(a\bar{x})$. Since $a\bar{x}$ is rational, there is at most a countable set of such spheres that contain the entire $\bar{X}$ therefore,

$$(4.5) \quad \bar{X} \subset \bigcup_{\bar{x} \in \bar{X}} B_{\bar{r}/2}(a\bar{x})$$

and from (2,4) this means $P(\bar{X}) = 0$. Therefore, increasing the size of training set can significantly improve of the classification results.

## 5. Experiments

In our experiments, we test the proposed methods on benchmark datasets. We use decision tree (C4.5) as base classifier using the 10-fold cross validation. To do so, the original dataset is divided to the test and training parts. Afterwards, Deep SMOTE and DA-SMOTE is used to over-sample the minority instances in training data as shown in Figure 2. Finally, a model is trained using the balanced training data by which the labels of test data are predicted. Precision, recall, F1 score and Area Under Curve (AUC) were averaged over 10-fold cross-validation runs for each of the data combinations. To reach a fair evaluation, all the metrics were averaged over 3 times of test on each dataset.

**5.1 Performance measures** Classifier performance metrics are typically evaluated by a confusion matrix, as shown in following table.

|  | Detected Pos | Detected Neg |
|---|---|---|
| **Actual Pos** | TP | FN |
| **Actual Neg** | FP | TN |

The rows are actual classes, and the columns are detected classes. TP (True Positive) is the number of correctly classified positive instances. FN (False Negative) is the number of incorrectly classified positive instances. FP (False Positive) is the number of incorrectly classified negative instances. TN (True Negative) is the number of correctly classified negative instances. The three performance measures are defined by formulae (1) through (3).

**Recall** = TP/(TP+ FN), **(1)**
**Precision** = TP/(TP+ FP), **(2)**
**F1** = (2* Recall * Precision) /(Recall+ Precision) **(3)**

**5.2 Datasets** In addition to the six benchmark datasets available in UCI data repository [1,2,3,4], we used Polish companies bankruptcy dataset [5]. The data contain 43405 instances with 64 features. The information has been collected from Emerging Markets Information Service in the period of 2000 to 2012, which is a database containing information on emerging markets around the world. These datasets are significantly different in terms of size and class proportions as summarized in Table 1.

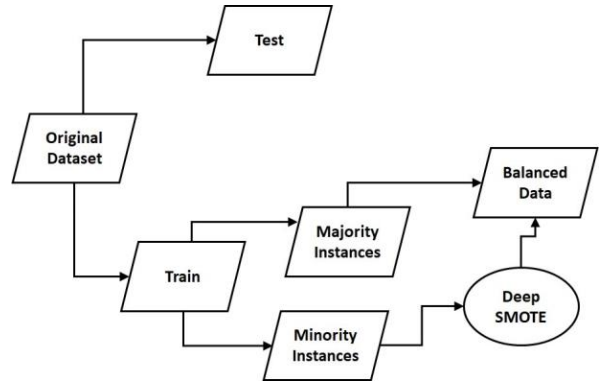

Figure 2: The Training - Test flowchart of Deep SMOTE

Table 2: Proposed architecture for generator and discriminator of DA-SMOTE corresponding to each dataset.

|  | Gen layers | Disc layers |
|---|---|---|
| WBC | 18,64,48,24,12,9 | 9,4,2,1 |
| Pima | 16,64,48,24,12,8 | 8,4,2,1 |
| Haberman | 6,64,48,24,6,3 | 3,12,8,6,2,1 |
| Ionosphere | 68,136,112,86,64,52,34 | 34,16,8,1 |
| Parkinson | 44,112,96,82,64,32,22 | 22,16,8,1 |
| Blood | 8,32,24,16,12,8,4 | 4,3,2,1 |
| Bankruptcy-1 | 128,512,256,128,100,82,64 | 64,48,24,16,8,1 |
| Bankruptcy-2 | 128,512,256,128,100,82,64 | 64,48,24,16,8,1 |
| Bankruptcy-3 | 128,512,256,128,100,82,64 | 64,48,24,16,8,1 |
| Bankruptcy-5 | 128,512,256,128,100,82,64 | 64,48,24,16,8,1 |

**5.3 Network Specification** To train DA-SMOTE and GAN we need to train two different networks: generator and discriminator. Table 2, Table 3 and Table 4 summarize the network specification related to proposed method and GAN for benchmark datasets.

Table 3: Proposed architecture for Deep SMOTE model corresponding to each dataset.

|  | Hidden Layers | Training Size |
|---|---|---|
| WBC | 48,32,16 | 12000 |
| Pima | 32,24,16 | 7000 |
| Haberman | 3 | 1500 |
| Ionosphere | 138,96,72,38 | 7000 |
| Parkinson | 40,36,32,28 | 1500 |
| Blood | 16,6 | 7000 |
| Bankruptcy Y-1 | 112,86,72 | 2000 |
| Bankruptcy Y-2 | 112,86,72 | 8000 |
| Bankruptcy Y-3 | 112,86,72 | 10000 |
| Bankruptcy Y-5 | 112,86,72 | 2000 |

Table 4: Proposed architecture for generator and discriminator of GAN corresponding to each dataset.

|  | Gen layers | Disc layers |
|---|---|---|
| WBC | 9,36,18,9 | 9,20,8,1 |
| Pima | 8,24,16,8 | 8,6,4,1 |
| Haberman | 3,9,6,3 | 3,10,8,1 |
| Ionosphere | 34,106,53,34 | 34,64,32,16,8,1 |
| Parkinson | 22,88,44,22 | 22,44,22,11,1 |
| Blood | 4, 16,8,4 | 4,16,8,1 |
| Bankruptcy Y-1 | 64, 16,8,64 | 64,32,16,8,1 |
| Bankruptcy Y-2 | 64, 16,8,64 | 64,32,16,8,1 |
| Bankruptcy Y-3 | 64, 16,8,64 | 64,32,16,8,1 |
| Bankruptcy Y-5 | 64, 16,8,64 | 64,32,16,8,1 |

## 6. Results

In this section, we report the comparison results of SMOTE, Deep SMOTE, DA-SMOTE, GAN, Borderline SMOTE and ADASYN as shown in Figure 3. We also applied paired t-test at 0.05 level for performing the significance tests between Deep SMOTE or DA-SMOTE and other methods. Colored square under each bar means that the proposed method significantly outperforms a baseline method. For example, blue square means DA-SMOTE and orange square means Deep SMOTE. Also, Table 5 tabulates the standard deviation results. Results can be categorized as follows:

• Deep SMOTE outperforms the SMOTE in terms of Precision, F1 and AUC: This type includes experimental results on five datasets including WBC, Pima, Ionosphere, Bankruptcy Year-1 and Bankruptcy Year-5 datasets.

• Deep SMOTE outperforms the SMOTE in terms of Precision and F1. This type includes experimental results on three datasets including Parkinson, Bankruptcy Y-2 and Bankruptcy Year-3 datasets.

• Deep SMOTE outperforms the SMOTE in terms of Precision and AUC. This type includes experimental results on Blood dataset.

• SMOTE outperforms Deep SMOTE in terms of all evaluation metrics. This type includes experimental results on Haberman dataset.

• Deep SMOTE outperforms both DA-SMOTE and SMOTE and all other methods in terms of all evaluated metrics. This type includes experimental results on Pima dataset.

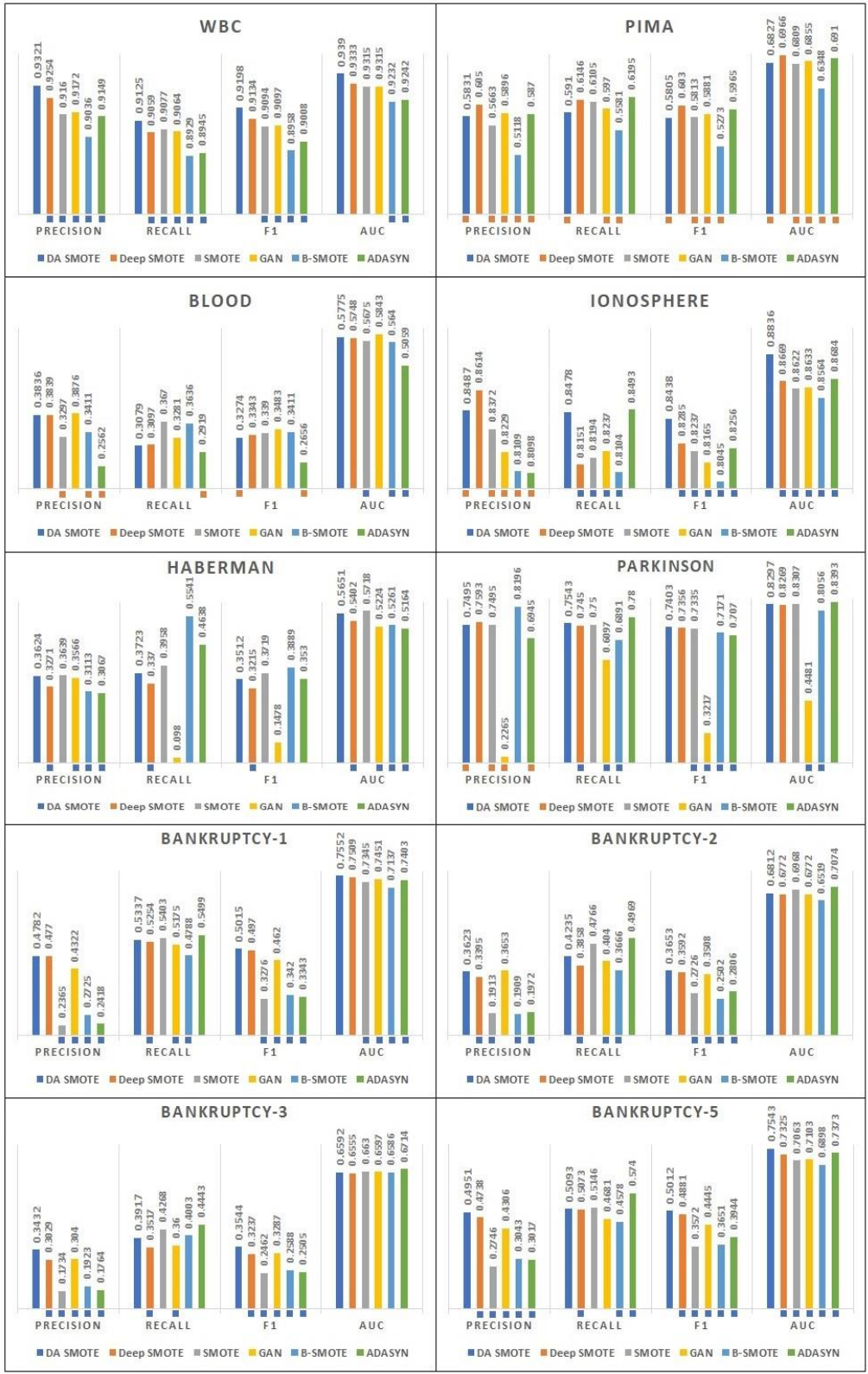

Figure 3: Experimental results based on 10 fold cross validation.

- DA-SMOTE outperforms both Deep SMOTE and SMOTE and all other methods in terms of all evaluated metrics. This type includes experimental results on WBC dataset.

- DA-SMOTE outperforms Deep SMOTE and SMOTE in terms of AUC. This type includes experimental results on five datasets including: WBC, Blood, Ionosphere and Bankruptcy Year 1,5.

- DA-SMOTE outperforms Deep SMOTE and SMOTE in terms of F1 score. This type includes experimental results on seven datasets including: WBC, Ionosphere, Parkinson and Bankruptcy Year- 1,2,3,5.

- In five datasets at least one of DA-SMOTE or Deep SMOTE outperforms other methods in terms of both F1 score and AUC.

- In eight datasets at least one of DA-SMOTE or Deep SMOTE outperforms other methods in terms of either F1 score or AUC.

**6.1 The curse of dimensionality** Haberman dataset is the only case that proposed methods fail to outperform SMOTE. One reason is the low dimensionality of Haberman dataset which is the lowest among the test cases. As the dimensionality increases, the Deep SMOTE and DA-SMOTE show better efficiency results. In general, adversarial over-sampling methods including DA-SMOTE and GAN can tolerate the curse of dimensionality better than other methods. Here, we summarize the comparison results between GAN and DA-SMOTE.

- In six datasets, DA-SMOTE outperforms GAN in terms of all metrics.

- In seven datasets, DA-SMOTE outperforms GAN in terms of F1 and AUC scores.

## 5 Conclusion

Imbalanced data classification has been extensively studied in the past decade. SMOTE as the most popular over-sampling technique and its variations still work based on $k-$ nearest neighbor and interpolation. To increase the efficiency and stability of SMOTE, we adopted its idea in deep learning architecture. The inputs of proposed regression model are two randomly chosen data points which are concatenated to form a double size vector. The model is trained to return the double sized input vector to the original dimension using a randomly interpolated data point between two corresponding concatenated data points. Concatenation of two randomly chosen data points to form a double size input vector helps us to adopt traditional SMOTE into deep learning architecture. Beyond that, it has a significant benefit: it increases the required training instances from $n$ to $C(n,2)$ which is vital to train a deep regression model. We also proposed the adversarial training of Deep SMOTE which is called Deep Adversarial SMOTE (DA-SMOTE). The advantage of DA-SMOTE versus Deep SMOTE is that, it's trained in unsupervised mode. According to our experimental results, Deep SMOTE and DA-SMOTE can enhance the classification results in majority of tested datasets.

Table 5: standard deviation results.

| | | F1 | AUC |
|---|---|---|---|
| **WBC** | DA SMOTE | 0.047 | 0.0305 |
| | Deep SMOTE | 0.0435 | 0.0341 |
| | SMOTE | 0.0465 | 0.0355 |
| **Pima** | DA SMOTE | 0.0798 | 0.0546 |
| | Deep SMOTE | 0.069 | 0.0458 |
| | SMOTE | 0.0785 | 0.0493 |
| **Blood** | DA SMOTE | 0.0984 | 0.0643 |
| | Deep SMOTE | 0.0978 | 0.0608 |
| | SMOTE | 0.0847 | 0.0668 |
| **Ionosphere** | DA SMOTE | 0.0587 | 0.0501 |
| | Deep SMOTE | 0.0636 | 0.0525 |
| | SMOTE | 0.0833 | 0.065 |
| **Haberman** | DA SMOTE | 0.1387 | 0.0854 |
| | Deep SMOTE | 0.1321 | 0.0835 |
| | SMOTE | 0.1183 | 0.079 |
| **Parkinson** | DA SMOTE | 0.1345 | 0.0933 |
| | Deep SMOTE | 0.1436 | 0.0957 |
| | SMOTE | 0.1811 | 0.123 |
| **Bankruptcy Y-1** | DA SMOTE | 0.0511 | 0.0422 |
| | Deep SMOTE | 0.0496 | 0.0419 |
| | SMOTE | 0.0775 | 0.0452 |
| **Bankruptcy Y-2** | DA SMOTE | 0.041 | 0.0367 |
| | Deep SMOTE | 0.0396 | 0.035 |
| | SMOTE | 0.0822 | 0.0499 |
| **Bankruptcy Y-3** | DA SMOTE | 0.0533 | 0.0466 |
| | Deep SMOTE | 0.0502 | 0.0443 |
| | SMOTE | 0.0716 | 0.0499 |
| **Bankruptcy Y-5** | DA SMOTE | 0.0734 | 0.0412 |
| | Deep SMOTE | 0.0784 | 0.0433 |
| | SMOTE | 0.0821 | 0.0481 |